\documentclass[sigconf,screen,nonacm]{acmart}
\usepackage{booktabs}  
\usepackage{multirow}  
\usepackage{graphicx}  
\usepackage{caption}
\usepackage{subcaption}
\usepackage{colortbl}  

\definecolor{GitHubLinkColor}{RGB}{25,115,200} 

\makeatletter
\renewcommand{\@parfont}{\bfseries}
\patchcmd{\@mkauthors@iii}
  {\unvbox\mktitle@bx\par\medskip\leavevmode}
  {\unvbox\mktitle@bx\par\medskip\vskip 6pt\leavevmode}
  {}{}
\makeatother

\AtBeginDocument{%
  }
\newcommand{\equalcontribmark}{\raisebox{0.05ex}{\scalebox{1.25}{*}}}

\title[Hallo-Live]{Hallo-Live: Real-Time Streaming Joint Audio-Video Avatar Generation with Asynchronous Dual-Stream and Human-Centric Preference Distillation}
\author[Li et al.]{%
  Chunyu Li\textsuperscript{1,2,}*
  \quad
  Jiaye Li\textsuperscript{2,}*
  \quad
  Ruiqiao Mei\textsuperscript{2}
  \quad
  Haoyuan Xia\textsuperscript{1,3}\\
  \quad
  Hao Zhu\textsuperscript{4}
  \quad
  Jingdong Wang\textsuperscript{5}\quad
  Siyu Zhu\textsuperscript{1,2,\textdagger}\\[10pt]
  \textsuperscript{1}Shanghai Innovation Institute\quad
  \textsuperscript{2}Fudan University\\
  \textsuperscript{3}University of Science and Technology of China\quad
  \textsuperscript{4}Nanjing University\quad
  \textsuperscript{5}Baidu
}
\thanks{\equalcontribmark\ Equal contribution.\\\textdagger\ Corresponding authors.}

\begin{document}
\begin{abstract}
Real-time text-driven joint audio-video avatar generation requires jointly synthesizing portrait video and speech with high fidelity and precise synchronization, yet existing audio-visual diffusion models remain too slow for interactive use and often degrade noticeably after aggressive acceleration. We present \textbf{Hallo-Live}, a streaming framework for joint audio-visual avatar generation that combines asynchronous dual-stream diffusion with human-centric preference-guided distillation. To reduce articulation lag in causal generation, we introduce \emph{Future-Expanding Attention}, which allows each video block to access synchronous audio together with a short horizon of future phonetic cues. To mitigate the quality loss of few-step distillation, we further propose \emph{Human-Centric Preference-Guided DMD} (HP-DMD), which reweights training samples using rewards from visual fidelity, speech naturalness, and audio-visual synchronization. On two NVIDIA H200 GPUs, Hallo-Live runs at 20.38 FPS with 0.94 seconds latency, yielding $16.0\times$ higher throughput and $99.3\times$ lower latency than the teacher model Ovi. Despite this speedup, it retains strong generation quality, reaching comparable VideoAlign overall score and Sync Confidence score while outperforming other accelerated baselines in the overall quality-efficiency trade-off. Qualitative results further show robust generalization across photorealistic, multi-speaker, and stylized scenarios. To the best of our knowledge, Hallo-Live is the first framework to combine streaming dual-stream diffusion with preference-guided distillation for real-time, text-driven audio-visual generation. Code and models are publicly available at \href{https://github.com/fudan-generative-vision/Hallo-Live}{\textcolor{GitHubLinkColor}{\textbf{\urlstyle{tt}\nolinkurl{https://github.com/fudan-generative-vision/Hallo-Live}}}}.
\end{abstract}


\begin{teaserfigure}
  \centering
  \vspace*{-15pt}
  \includegraphics[width=0.96\textwidth]{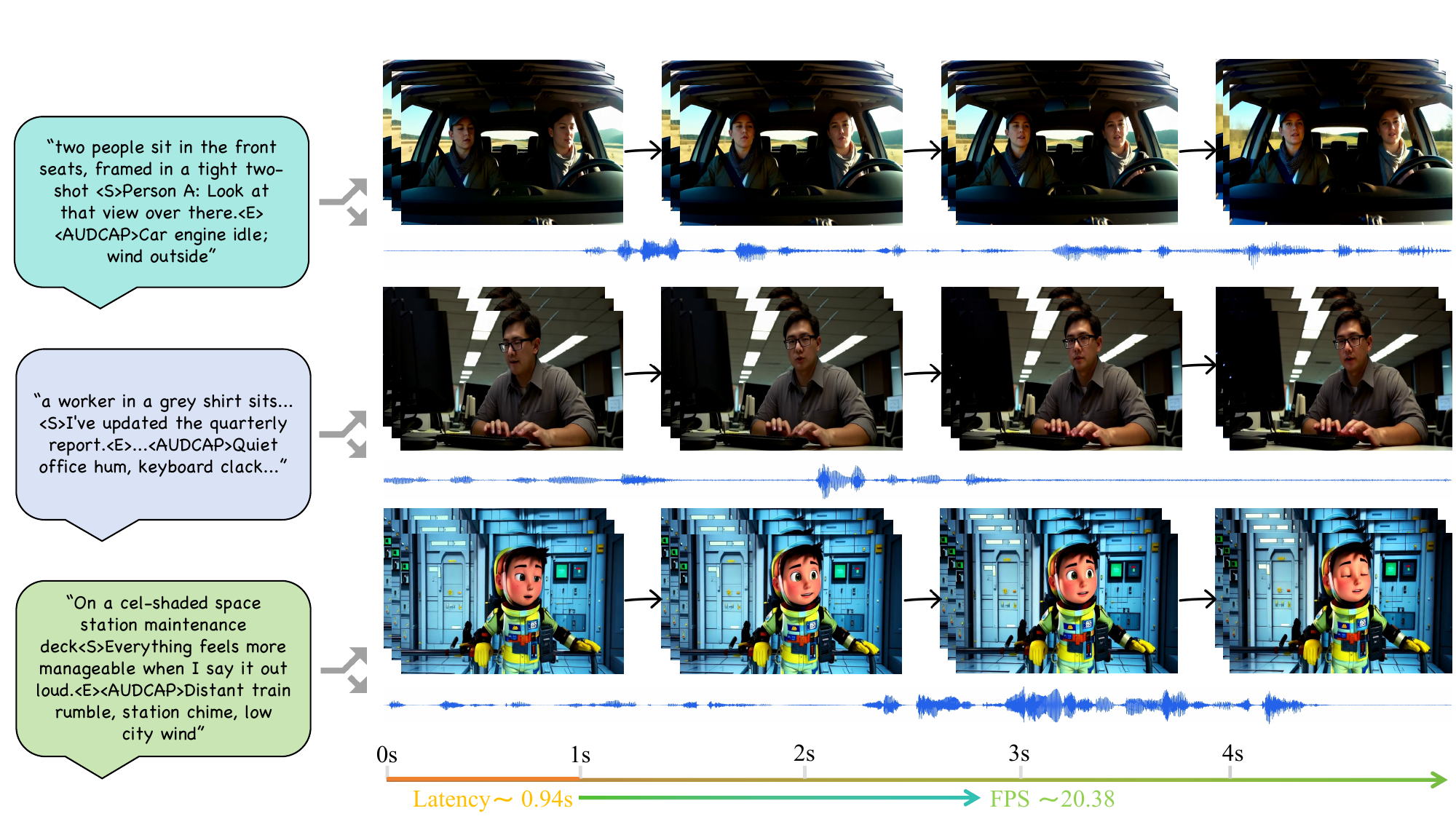}
  \caption{Our method enables real-time streaming text-driven joint audio-video avatar generation. On two NVIDIA H200 GPUs, Hallo-Live reaches \textcolor{red}{20.38 FPS} with \textcolor{red}{0.94s latency} while preserving strong lip-sync accuracy and visual fidelity. The examples above show robust generalization across diverse scenarios, including multi-speaker interactions, photorealistic portraits, and stylized cartoon characters.}
  \label{fig:teaser}
  \Description{Overview examples of Hallo-Live showing text-driven audio-visual avatar generation across multi-speaker, photorealistic, and stylized cases, together with real-time throughput and latency results.}
\end{teaserfigure}

\maketitle
\section{Introduction}
Text-driven joint audio-video avatar generation aims to synthesize coherent avatar video and speech from natural-language prompts. This setting inherits prompt conditioning from large text-to-text backbones such as T5 \cite{2020t5} and benefits from transformer-based diffusion architectures rooted in self-attention and rotary positional encoding \cite{vaswani2017attention,su2024roformer,peebles2023scalable}. Recent advances in latent diffusion and multimodal generation \cite{rombach2022high,wang2024emu3,wan2025wan,zhang2026foleycrafter}, together with earlier joint audio-visual generation efforts \cite{ruan2023mm,liu2025javisdit,low2025ovi}, have pushed this task forward, and Ovi \cite{low2025ovi} shows that a dual-stream diffusion architecture can produce high-quality synchronized audio-visual outputs.

However, real-time audio-video avatar generation remains difficult. Early audio-video diffusion models are too slow for interactive use \cite{hacohen2026ltx,low2025ovi,team2026mova}. Although recent streaming-oriented method OmniForcing \cite{su2026omniforcing} utilizes Self-forcing technique \cite{huang2025self} to transform a bidirectional joint audio-video model into a causal model, it has two major issues. First, causal dual-stream inference makes it hard to preserve the short-horizon future audio required for natural lip motion. Second, aggressive distillation often leads to mean-seeking artifacts that degrade visual fidelity, speech quality, and cross-modal consistency.

In this work, we present \textbf{Hallo-Live}, a real-time framework for text-driven joint audio-video avatar generation. Our first component is an asynchronous dual-stream diffusion architecture tailored for streaming inference. We observe that realistic facial articulation depends on short-horizon future phonetic cues, whereas standard causal masking exposes the video stream only to current and past audio. To address this mismatch, we introduce Future-Expanding Attention, which allows each video block to attend to synchronous audio together with a short look-ahead region. During causal inference, we concatenate extra future audio noise to the current audio noise input so that the audio stream directly denoises a short future span, enabling anticipatory lip motion without breaking streaming causality.

Our second component is Human-Centric Preference-Guided DMD (HP-DMD), which reduces the quality loss caused by aggressive acceleration. Instead of treating all teacher samples equally, HP-DMD reweights distillation updates using reward signals from SyncNet \cite{chung2016out}, VideoAlign \cite{liu2025improving}, and AudioBox \cite{tjandra2025meta}. This biases learning toward samples with better synchronization, stronger visual fidelity, and more natural speech, yielding a more favorable quality-efficiency trade-off than vanilla DMD.

To our knowledge, 
Hallo-Live is the first framework to unify streaming dual-stream diffusion with preference-guided distillation for real-time, 
text-driven audio-visual generation. 
Benchmarked on two NVIDIA H200 GPUs, Hallo-Live achieves 20.38 FPS with a 0.94-second latency, 
representing a 16.0$\times$ increase in throughput and a 99.3$\times$ reduction in latency relative to the teacher model Ovi. 
Despite this significant acceleration, 
the framework maintains high generative fidelity, delivering synchronization and visual alignment comparable to the teacher while surpassing previous accelerated baselines in the overall quality-efficiency trade-off. 
Finally, qualitative results demonstrate that Hallo-Live is highly versatile, achieving robust generalization across diverse photorealistic, multi-speaker, and stylized scenarios.

\begin{figure*}[t]
    \centering
    \includegraphics[width=\linewidth]{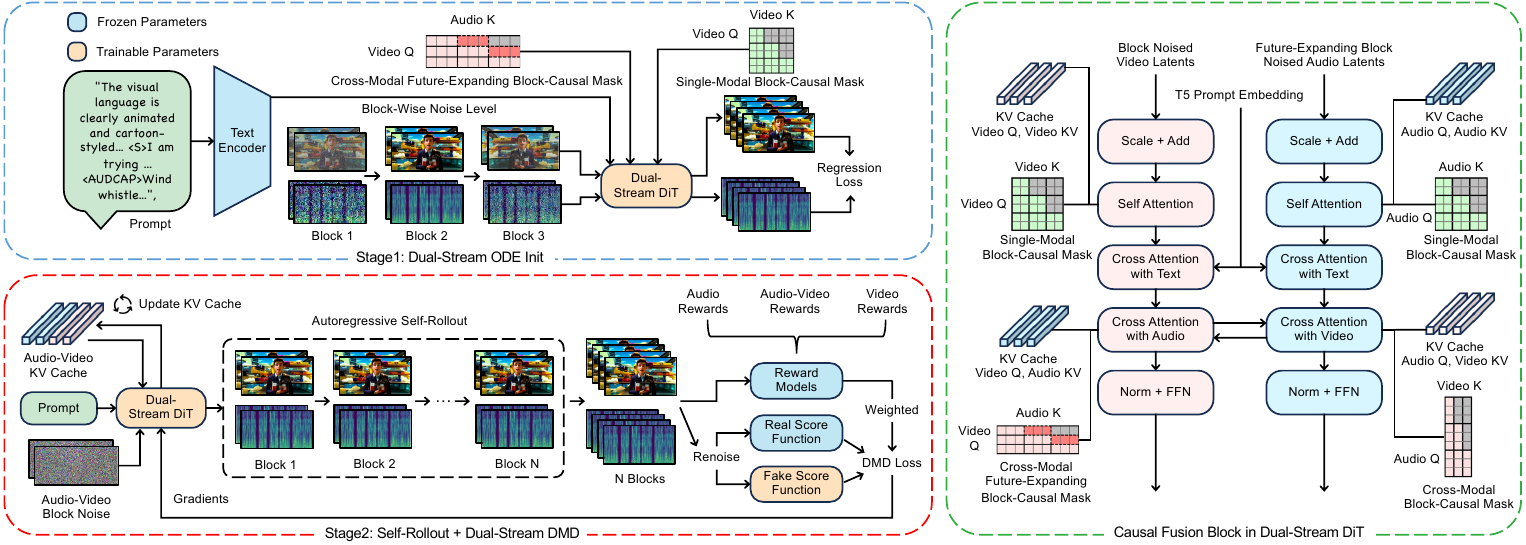}
    \caption{Overview of Hallo-Live. Top left: Stage I adapts a pretrained dual-stream DiT to the streaming setting using cross-modal future-expanding block-causal mask. Bottom left: Stage II performs autoregressive self-rollout with an audio-video KV cache and optimizes the generated trajectory with reward-weighted dual-stream DMD. Right: Each causal fusion block in the dual-stream DiT consists of single-modal block-causal self-attention, text cross-attention, and cross-modal attention between the video and audio streams, where the block-causal masks are utilized in Stage I ODE initialization, and KV cache is maintained for Stage II self-rollout and streaming inference.}
    \label{fig:network}
\end{figure*}

\section{Related Work}
\noindentparagraph{Portrait Animation and Talking Avatars.}
Traditional speech-driven portrait animation focuses on mapping acoustic features to facial dynamics. 
Early benchmarks, such as Wav2Lip~\cite{prajwal2020wav2lip} and SadTalker~\cite{zhang2023sadtalker}, 
prioritized lip synchronization and 3D structural consistency. 
Recent diffusion-based frameworks including EMO~\cite{tian2024emo}, VASA-1~\cite{xu2024vasa1}, LatentSync~\cite{li2024latentsync}, Hallo series~\cite{xu2024hallo,cui2024hallo2,cui2025hallo3,cui2025hallo4} and other works \cite{ma2023dreamtalk,chen2024echomimic,zhu2025infp,jiang2024loopy,zhang2024musetalk,mukhopadhyay2024diff2lip,bigioi2024speech,ji2025sonic,peng2025omnisync} have significantly elevated visual fidelity and motion expressiveness. 
While recent systems like Teller~\cite{zhen2025teller} and OmniAvatar~\cite{gan2025omniavatar} further explore streaming inference and cinematic control, they remain essentially audio-to-video (A2V) models that assume the existence of a driving audio signal.


\noindentparagraph{Joint Audio-Video Generation.}
Beyond traditional audio-to-video mapping, recent research has shifted toward modeling audio and video as a unified generative process. 
Early joint-generation systems such as MM-Diffusion~\cite{ruan2023mm} uses sequential multi-modal U-Net \cite{ronneberger2015u} for a joint denoising process. JavisDiT~\cite{liu2025javisdit} introduces hierarchical spatio-temporal priors within a joint diffusion transformer \cite{peebles2023scalable} to improve synchronization, while Ovi~\cite{low2025ovi} employs a twin-backbone architecture with bidirectional cross-modal fusion for text-driven synthesis. UniVerse-1~\cite{wang2025universe} leverages a stitching of experts (SoE) approach to fully leverage the capabilities of foundation models across modalities. DaVinci-MagiHuman~\cite{chern2026speed} utilizes single-stream simplification to avoid the complexity of multi-stream. MOVA~\cite{team2026mova} employs a Mixture-of-Experts (MoE) architecture~\cite{shazeer2017outrageously} to further scale up the model capacity. LTX-2~\cite{hacohen2026ltx} utilizes the modality-aware classifier-free guidance mechanism for improved audio-video alignment. OmniForcing~\cite{su2026omniforcing} uses DMD distillation \cite{yin2024one}, and autoregressive self-rollout training strategies \cite{huang2025self} to equip audio-video models with streaming generation capabilities.



\noindentparagraph{Distribution Matching Distillation and Preference Alignment.}
DMD~\cite{yin2024one} provides a robust framework for accelerating diffusion models by aligning the student’s generative distribution with a pre-trained teacher's manifold. 
The following works, such as DMD2~\cite{yin2024improved}, 
incorporate adversarial loss to enhance image sharpness, 
while $f$-distill~\cite{xu2025one} and TDM~\cite{luo2025learning} optimize distributional coverage and trajectory alignment.
Parallel to distillation, preference alignment via reinforcement learning has emerged as a key paradigm for steering generative outputs toward human aesthetics. 
While DDPO~\cite{black2023ddpo} and human-feedback-guided diffusion~\cite{lee2023aligning} demonstrate success in text-to-image tasks, 
VideoAlign~\cite{liu2025improving} extends this to the temporal domain using multi-dimensional reward models. 
Recently, DMDR~\cite{jiang2025distribution} and RewardForcing~\cite{lu2025reward} combine reinforcement learning with distribution matching distillation, suggesting that reward-aware distillation can transcend pure teacher imitation.

Our work bridges these paradigms within the more rigorous dual-modal setting. 
Unlike single-modal acceleration, our approach must simultaneously preserve human-centric visual fidelity, speech naturalness, and, crucially, timestamp-level audio-video synchronization. 
To this end, we introduce a multi-modal preference-aware reweighting mechanism specifically engineered for the unique constraints of avatar audio-video distillation.

\section{Method}

\subsection{Overview}
\noindentparagraph{Causal Dual-Stream Audio-Video Diffusion.}
As illustrated in Figure~\ref{fig:network}, Hallo-Live is built on a text-conditioned dual-stream DiT that jointly denoises block-wise video and audio latents within a unified audio-video generation process \cite{low2025ovi}. 
The backbone contains parallel video and audio branches connected by causal fusion blocks: each branch applies single-modal block-causal self-attention, injects the text condition, and then exchanges information through cross-modal attention between the two streams.
On top of this causal dual-stream backbone, we adopt a two-stage training pipeline: Stage~I initializes the streaming student from a pretrained Ovi teacher under the new masking pattern, and Stage~II further performs autoregressive self-rollout with dual-stream DMD to improve audio-video fidelity and synchronization.

Nevertheless, adapting this expressive dual-stream architecture to a real-time setting with DMD remains challenging, mainly due to two technical bottlenecks:

\noindentparagraph{1) Limited Audio Context in Dual-Stream Causal Inference.} 
Real-time acceleration requires the dual-stream model to operate under a causal streaming constraint, so the cross-modal interaction between audio and video must also follow block-wise causality. 
In practice, realistic facial articulation and upper-body motion, especially lip movements, depend not only on the current audio segment but also on short-horizon upcoming phonetic cues. 
However, under standard causal or strictly synchronous windowing, the dual-stream inference process still exposes the video stream to only the current and past audio blocks, while informative near-future speech context remains inaccessible. This limited audio context results in delayed or imprecise articulations and degraded lip-sync quality.

\noindentparagraph{2) Distillation-Induced Human-Centric Degradation.}
While DMD effectively accelerates inference by aligning student distributions with a pre-trained teacher manifold, vanilla distillation often leads to ``mean-seeking'' artifacts, and degradation in human-centric metrics.
In the context of avatar animation, 
this manifests as a loss of facial and body fine-grained visual textures, 
rigid and robotic speech prosody, 
and accumulated temporal drift in audio-video alignment.

\subsection{Asynchronous Dual-Stream Diffusion}
\begin{figure}[t]
    \centering
    \includegraphics[width=\linewidth]{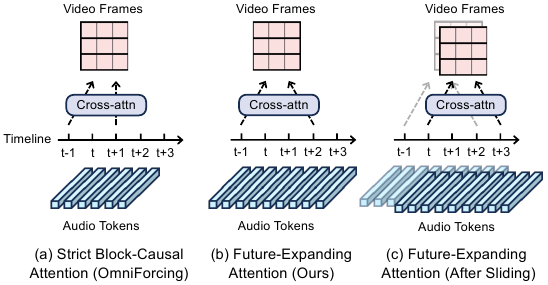}
    \caption{Comparison of attention mechanisms. (a) The Strict Block-Causal Attention uses synchronous alignment, where the current video block only attends to the current audio block and past context. (b) Our Future-Expanding Attention expands the audio receptive field to include the look-ahead blocks, enabling anticipatory audio-visual synchronization. (c) After the window slides forward, the overlapped audio context is retained and the provisional future block is refreshed.}
    \label{fig:attention}
\end{figure}

\noindentparagraph{Bottleneck of Strict Block-Causal Attention.}
To achieve streaming inference, 
a common baseline (shown in Figure~\ref{fig:attention}~(a)) is the strict block-causal attention. 
In this configuration, 
the video stream $f_v$ and audio stream $f_a$ are partitioned into temporal blocks of duration $\Delta$ (e.g., 1s). 
Let $\mathcal{B}_t = \{V_t, A_t\}$ denote the $t$-th latent block pair. 

While this ensures temporal consistency, 
it imposes a strict block-causal receptive field that prevents the video stream from accessing near-future phonetic context. 
Human speech involves significant co-articulation where lip movements often precede acoustic onset.
At the same time, expressions and body movements also rely on the understanding of subsequent speech audio within a larger scope of events.
Consequently, a strictly causal receptive field limits the model's capacity for phonetic anticipation, 
leading to visible lag and reduced lip-sync precision during causal inference.

\noindentparagraph{Future-Expanding Attention.}
To resolve this, we propose Future-Expanding Attention (shown in Figure~\ref{fig:attention}~(b)). 
Unlike the strict block-causal attention that forces a symmetric temporal boundary, 
our approach asymmetrically expands the audio context relative to the video query. 
At any inference step $t$, 
the video branch maintains a localized focus on the current block, 
while the audio branch provides a wider temporal support consisting of historical, 
synchronous, 
and look-ahead segments.

Formally, for the $t$-th interval $[t\Delta, (t+1)\Delta]$, 
we define the operative windows as:
\begin{equation}
\mathcal{W}_t^{v} = \{V_t\}, \quad \mathcal{W}_t^{a} = \{\hat{A}_{t-1}, A_t, \tilde{A}_{t+1}\}
\end{equation}
where $\hat{A}_{t-1}$ is the committed audio from the previous step and $\tilde{A}_{t+1}$ is a temporary look-ahead block obtained from an expanded audio-noise input. 
When computing cross-attention for the video stream, the video tokens $Q_t^v$ query an expanded key range:
\begin{equation}
H_t^{v} = \text{CrossAttn}\left(Q_t^{v}, \text{KV}(\hat{A}_{t-1} \oplus A_t \oplus \tilde{A}_{t+1})\right)
\end{equation}
This future-expanding configuration grants the video stream a ``preview'' of upcoming phonetic dynamics,
effectively modeling the natural lead-time required for realistic facial articulation.

\noindentparagraph{Asynchronous Dual-Stream Diffusion.}
We realize asynchronous dual-stream diffusion by advancing the video and audio branches with different temporal scopes at each streaming step. At step $t$, the video branch only denoises the current video-block noise $\mathbf{z}_{t}^{v}$ and commits $V_t$, whereas the audio branch receives an expanded noise input
\begin{equation}
    \mathbf{z}_{t}^{a,+} = \mathbf{z}_{t}^{a} \oplus \mathbf{z}_{t+1}^{a},
\end{equation}
where $\mathbf{z}_{t}^{a}$ denotes the noise for the current audio block and the concatenated term represents extra future audio-noise frames. The two streams therefore evolve under different temporal states: the video stream remains on a single committed block, while the audio stream simultaneously models the committed current block and a provisional future block. The joint denoising at step $t$ thus produces $(V_t, A_t, \tilde{A}_{t+1})$, where $A_t$ is committed as the current audio block and $\tilde{A}_{t+1}$ is retained only as look-ahead context for cross-modal interaction. In practice, the expanded audio-noise input can include several future frames; we write one additional block here for simplicity.

After the window slides by one block, the schedule becomes
\begin{equation}
    \mathcal{W}_{t+1}^{v} = \{V_{t+1}\}, \qquad
    \mathcal{W}_{t+1}^{a} = \{\hat{A}_{t}, A_{t+1}, \tilde{A}_{t+2}\}.
\end{equation}
The temporary block $\tilde{A}_{t+1}$ is never committed directly. Once the window slides, the video stream advances to $V_{t+1}$, while the audio stream shifts its wider state forward and denoises the new expanded input to produce $A_{t+1}$ together with a refreshed look-ahead block $\tilde{A}_{t+2}$. The earlier provisional block therefore serves only as transient conditioning and is overwritten before commitment. Consequently, the model can provide anticipatory phonetic cues to the video stream without accumulating speculative audio errors, introducing only one-block look-ahead latency while improving lip anticipation and timestamp-level synchronization.

\begin{figure}[t]
    \centering
    \includegraphics[width=0.6\linewidth]{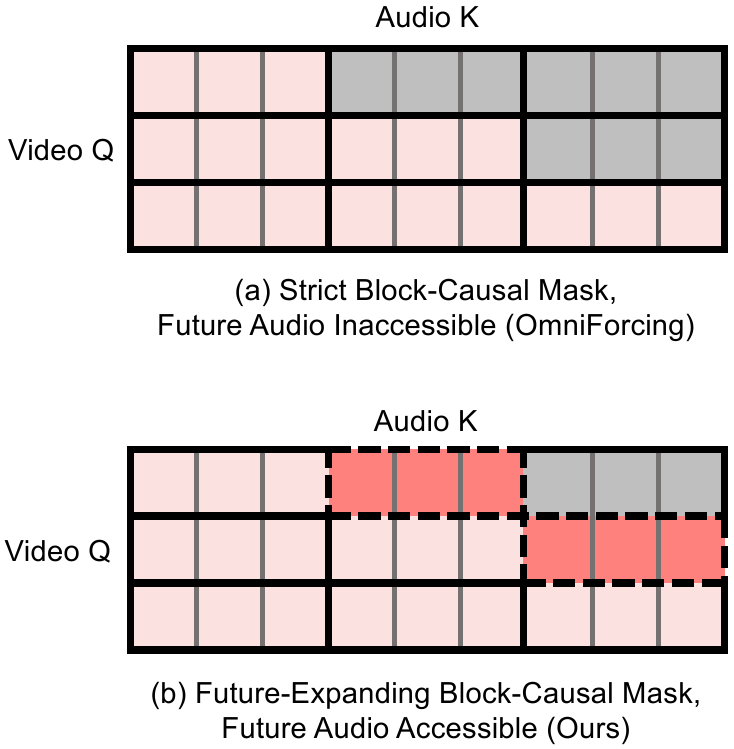}
    \caption{Cross-modal causal masks from video queries to audio keys. (a) In the Strict Block-Causal mask, future audio positions remain inaccessible. (b) Our Future-Expanding Block-Causal mask selectively reveals a short look-ahead audio region.}
    \label{fig:causal_mask}
\end{figure}

\noindentparagraph{Future-Expanding Block-Causal Mask.}
The asynchronous update rule above defines the streaming inference schedule, but the model must also be trained under the same visibility pattern during Stage I ODE initialization. To this end, we introduce a cross-modal mask $M^{v\leftarrow a}$ from video queries to audio keys. Let $t$ denote the index of a query video frame and $s$ the index of a key audio token. Since one video frame is temporally aligned with $r=5$ audio tokens, the Future-Expanding Block-Causal Mask with a look-ahead window $W$ measured in video frames is defined as
\begin{equation}
M_{t,s}^{v\leftarrow a}(W)=
\begin{cases}
1, & s \le r(t+W),\\
0, & s > r(t+W),
\end{cases}
\end{equation}
where $W=1$ gives the one-frame look-ahead window visualized in Figure~\ref{fig:causal_mask}. The strict block-causal mask is recovered by setting $W=0$, in which case video frame $V_t$ can attend only to past audio tokens and the five synchronous tokens aligned with that frame. By contrast, our mask additionally reveals the next five audio tokens aligned with $V_{t+1}$ while keeping all later future audio positions inaccessible. This future-expanding visibility pattern is the training-time realization of Future-Expanding Attention, teaching the student to rely on the same limited future phonetic context that will be available during streaming inference and DMD training, thereby improving anticipatory audio-visual synchronization without introducing unrestricted future leakage.

\subsection{Human-Centric Preference-Guided DMD}
To mitigate the performance degradation and ``mean-seeking'' artifacts typically associated with vanilla distillation, 
we propose human-centric preference-guided DMD (HP-DMD). 
Unlike standard DMD, 
which forces the student model to replicate the teacher's entire output manifold,
HP-DMD integrates fine-grained reward modeling and dynamic importance sampling,
steering the student's generative distribution toward human-centric metrics of avatar audio-visual generation,
such as human visual fidelity, acoustic naturalness, and audio-visual synchronization.
This mechanism effectively allows the student to surpass the average performance ceiling of the teacher model.

\noindentparagraph{Multi-Modal Reward Modeling.}
Given a batch of $B$ text prompts $\{y_1, y_2, \dots, y_B\}$, the student model $S_\theta$ generates audio-visual samples $x_i = S_\theta(y_i)$. We then evaluate each sample with three reward models:
\begin{itemize}
    \item Visual fidelity ($R_{v}$): VideoAlign \cite{liu2025improving}, which measures visual quality, motion quality, and text alignment.
    \item Acoustic naturalness ($R_{a}$): AudioBox \cite{tjandra2025meta}, which evaluates the perceptual quality of synthesized speech.
    \item Audio-visual synchronization ($R_{s}$): a SyncNet-based score \cite{chung2016out}, which measures lip-audio alignment.
\end{itemize}
The reward of metric $k$ for sample $x_i$ is
\begin{equation}
    R_{i,k} = \text{Metric}_k(x_i, y_i), \quad k \in \{1, \dots, K\}.
\end{equation}

\noindentparagraph{Batch-wise Standardization and Reweighting.}
The raw rewards have different scales and vary with prompt difficulty, so we standardize them within each batch before combining them:
\begin{equation}
    z_{i,k} = \frac{R_{i,k} - \mu_k}{\sigma_k + \epsilon}
\end{equation}
where $\mu_k$ and $\sigma_k$ are the mean and standard deviation of metric $k$ over the batch:
\begin{equation}
    \mu_k = \frac{1}{B} \sum_{j=1}^B R_{j,k}, \quad \sigma_k = \sqrt{\frac{1}{B} \sum_{j=1}^B (R_{j,k} - \mu_k)^2}
\end{equation}

We then aggregate the standardized rewards into a sample weight
\begin{equation}
\label{eq:weights}
w_i = \exp\left( \sum_{k=1}^K \beta_k z_{i,k} \right)
\end{equation}
where $\beta_k$ controls the contribution of each modality. Samples with better relative reward therefore contribute larger gradients during distillation.

\noindentparagraph{Distribution Refinement.}
The final HP-DMD objective is the weighted DMD loss
\begin{equation}
    \mathcal{L}_{final}(\theta, x_i) = w_i \cdot \mathcal{L}_{dmd}(x_i)
\end{equation}
which can be interpreted as fitting a reward-tilted target distribution $p^* \propto p_T \cdot \exp(R)$ rather than the original teacher distribution $p_T$. In practice, this shifts optimization toward regions of the teacher manifold with higher visual fidelity, better speech quality, and stronger synchronization, yielding a better quality-efficiency trade-off after distillation.

\subsection{Architecture and Training Pipeline}

\noindentparagraph{Causal Fusion Block.}
As shown in Figure~\ref{fig:network}, Hallo-Live is initialized from a pretrained Ovi model and replaces the original fully bidirectional temporal interaction with a causal fusion block tailored to streaming generation. In each dual-stream DiT block, the video and audio latents interact through single-modal self-attention, text cross-attention, and cross-modal attention between the two streams. During Stage~I ODE initialization, these interactions are adapted to the streaming setting using single-modal block-causal masks together with a Future-Expanding cross-modal mask. During Stage~II self-rollout and inference stage, the model maintains a rolling audio-video KV cache over committed history to support efficient causal generation.

\noindentparagraph{Stage I: Dual-Stream ODE Initialization.}
We first adapt the pretrained backbone to the causal masking pattern in Figure~\ref{fig:network} without performing long-horizon autoregressive rollout. Let $v_{\phi}$ denote the frozen Ovi teacher and $v_{\theta}$ the student equipped with the single-modal block-causal mask and the cross-modal Future-Expanding mask. Let $y$ denote the text condition encoded from the input prompt. For a noisy joint latent $\mathbf{x}_t = [V_t, A_t]$ at flow time $t$, we regress the student prediction to the teacher trajectory:
\begin{equation}
\begin{aligned}
\mathcal{L}_{\mathrm{init}}
&= \mathbb{E}_{t, \mathbf{x}_t} \Big[
\lambda_v \left\| v_{\theta}^{v}(\mathbf{x}_t, y) - v_{\phi}^{v}(\mathbf{x}_t, y) \right\|_2^2 \\
&\quad + \lambda_a \left\| v_{\theta}^{a}(\mathbf{x}_t, y) - v_{\phi}^{a}(\mathbf{x}_t, y) \right\|_2^2
\Big].
\end{aligned}
\end{equation}
This stage transfers the teacher's joint audio-video denoising capabilities to the student, allowing the causal fusion blocks to inherit the pretrained prior before exposure to autoregressive errors.

\noindentparagraph{Stage II: Self-Rollout and Dual-Stream DMD.}
After initialization, the student autoregressively generates a sequence of audio-video blocks $\hat{\mathcal{B}}_1, \ldots, \hat{\mathcal{B}}_K$ under the same causal fusion mechanism used at inference time, while the audio-video KV cache is updated online from the committed history. This self-rollout stage repeatedly exposes the student to its own prediction history, enabling dual-stream DMD to correct accumulated drift in visual fidelity, speech quality, and audio-visual synchronization. Let $\hat{\mathbf{V}}$ and $\hat{\mathbf{A}}$ denote the rolled-out video and audio latents. Rather than applying a single monolithic DMD loss to the concatenated audio-video sample, we compute modality-specific DMD gradients for the two streams. Given renoised latents $(\tilde{\mathbf{V}}_{\tau}, \tilde{\mathbf{A}}_{\tau})$ at timestep $\tau$, the fake and real score networks produce video and audio predictions, yielding normalized DMD gradients
\begin{equation}
\begin{aligned}
g^{v} &=
\frac{s_{\mathrm{fake}}^{v}(\tilde{\mathbf{V}}_{\tau}, \tilde{\mathbf{A}}_{\tau}, y)
- s_{\mathrm{real}}^{v}(\tilde{\mathbf{V}}_{\tau}, \tilde{\mathbf{A}}_{\tau}, y)}
{\mathcal{N}_{v}},\\
g^{a} &=
\frac{s_{\mathrm{fake}}^{a}(\tilde{\mathbf{V}}_{\tau}, \tilde{\mathbf{A}}_{\tau}, y)
- s_{\mathrm{real}}^{a}(\tilde{\mathbf{V}}_{\tau}, \tilde{\mathbf{A}}_{\tau}, y)}
{\mathcal{N}_{a}},
\end{aligned}
\end{equation}
where $\mathcal{N}_{v}$ and $\mathcal{N}_{a}$ are modality-specific normalization factors. We then form separate DMD surrogate losses for video and audio:
\begin{equation}
\begin{aligned}
\mathcal{L}_{\mathrm{dmd}}^{v}
&= \frac{r_v}{2}
\left\|\hat{\mathbf{V}} -
\mathrm{sg}\left(\hat{\mathbf{V}} - g^{v}\right)\right\|_2^2,\\
\mathcal{L}_{\mathrm{dmd}}^{a}
&= \frac{r_a}{2}
\left\|\hat{\mathbf{A}} -
\mathrm{sg}\left(\hat{\mathbf{A}} - g^{a}\right)\right\|_2^2,
\end{aligned}
\end{equation}
where $\mathrm{sg}(\cdot)$ denotes stop-gradient. The final Stage~II objective is the weighted sum of the two stream-specific losses:
\begin{equation}
\begin{aligned}
\mathcal{L}_{\mathrm{rollout}}
&= \gamma_v \mathcal{L}_{\mathrm{dmd}}^{v}
+ \gamma_a \mathcal{L}_{\mathrm{dmd}}^{a}.
\end{aligned}
\end{equation}
Here $\gamma_v$ and $\gamma_a$ balance the two modalities. The reward scales $r_v$ and $r_a$ are computed from the decoded rollout: VideoAlign and SyncNet modulate the video DMD term, while AudioBox and SyncNet modulate the audio DMD term.

\begin{figure*}[t]
    \centering
    \includegraphics[width=\textwidth]{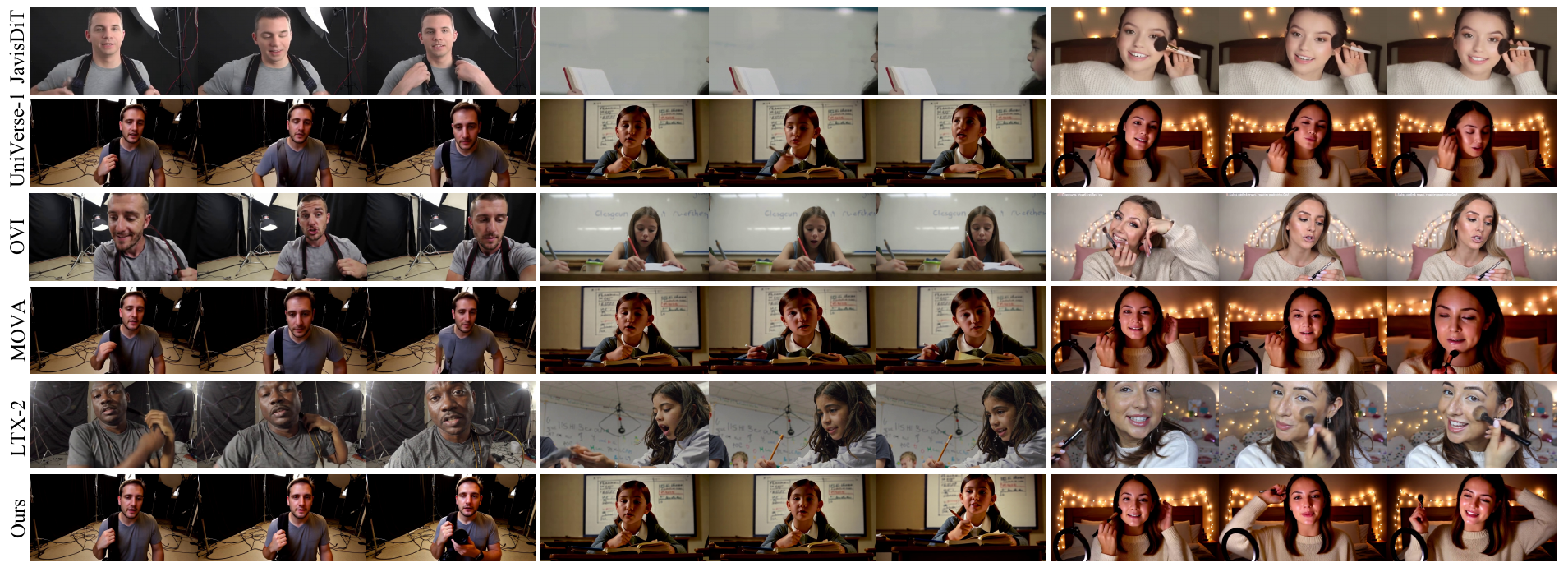}
    \caption{Comparison with state-of-the-art methods (Ovi \cite{low2025ovi}, UniVerse-1 \cite{wang2025universe}, JavisDiT \cite{liu2025javisdit}, MOVA \cite{team2026mova},LTX-2 \cite{hacohen2026ltx}). Our method achieves competitive or superior performance across multiple metrics, particularly after preference distillation.}
    \label{fig:main_results}
\end{figure*}
\begin{figure*}[t]
    \centering
    \includegraphics[width=\textwidth]{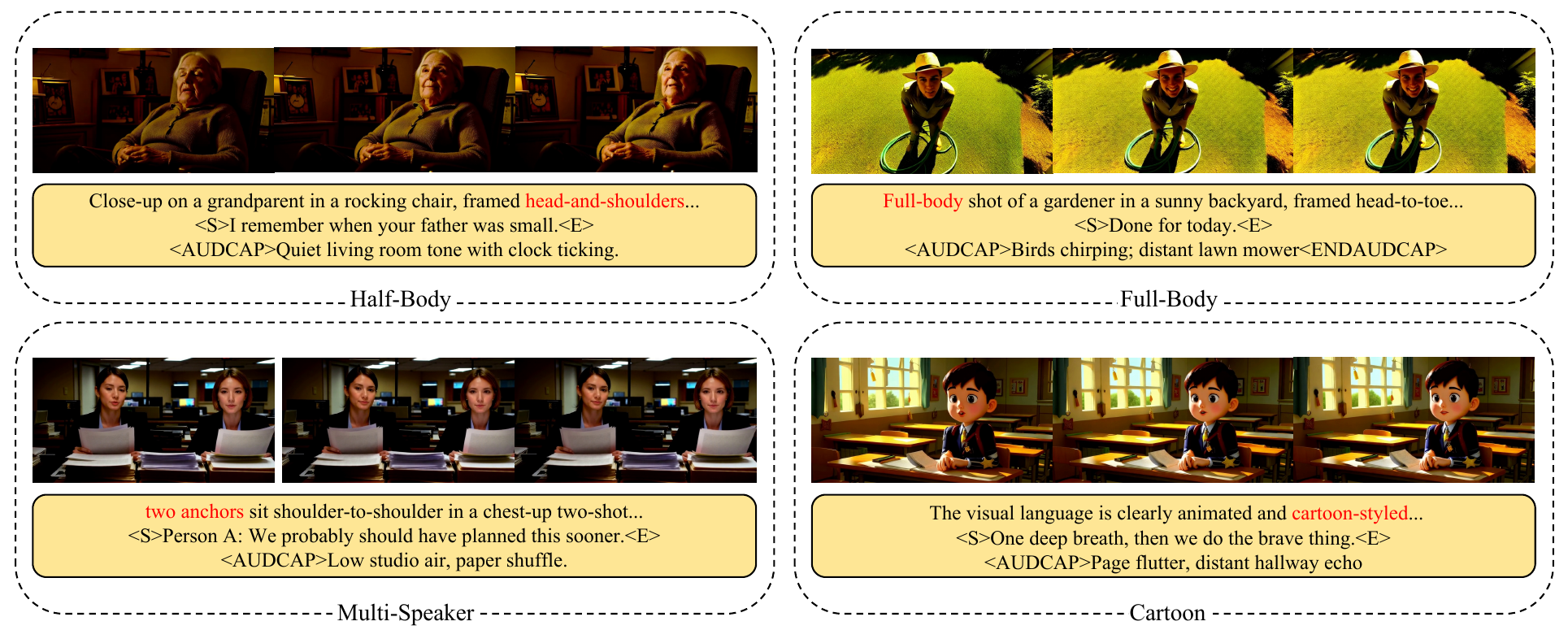}
    \caption{Generation results with different prompts. The figure showcases diverse generation capabilities, ranging from specific spatial compositions like Half-Body (top-left) and Full-Body (top-right) framing, to complex Multi-Speaker dynamics (bottom-left) and Cartoon stylization (bottom-right). Our method accurately captures prompt-specified attributes. }
    \label{fig:prompt}
\end{figure*}
\section{Experiments}
\noindentparagraph{Implementation Details.}
\begin{table*}[t]
\centering
\scriptsize
\setlength{\tabcolsep}{4pt}
\renewcommand{\arraystretch}{0.92}
\resizebox{\textwidth}{!}{%
\begin{tabular}{l c c c c c c c c c c c c c c c}
\toprule
\multirow{2}{*}{Method} & 
\multirow{2}{*}{\shortstack{Throughput \\(FPS)$\uparrow$}} & 
\multirow{2}{*}{\shortstack{Latency \\(s)$\downarrow$}} & 
\multicolumn{4}{c}{VideoAlign} & 
\multirow{2}{*}{Sync-C$\uparrow$} & 
\multicolumn{3}{c}{AudioBox} & 
\multicolumn{2}{c}{TTS} &
\multicolumn{3}{c}{Human Fid.} \\
\cmidrule(lr){4-7} \cmidrule(lr){9-11} \cmidrule(lr){12-13} \cmidrule(lr){14-16}
 & & & VQ$\uparrow$ & MQ$\uparrow$ & TA$\uparrow$ & Overall$\uparrow$ & & 
CE$\uparrow$ & CU$\uparrow$ & PQ$\uparrow$ & CLAP$\uparrow$ & WER$\downarrow$ & Anat.$\uparrow$ & Clo.$\uparrow$ & Id.$\uparrow$ \\
\midrule
JavisDiT & 2.15 & 24.40 & -0.18 & 0.55 & 0.66 & 1.03 & 1.12 & 3.64 & 4.28 & 5.51 & 0.19 & 0.88 & 0.90 & 0.94 & 0.93\\
\midrule
UniVerse-1 & 0.64 & 187.76 & -0.20 & 0.26 & 1.14 & 1.20 & 1.46 & 4.02 & 4.30 & 4.75 & 0.18 & 0.07 & 0.78 & 0.96 & 0.82\\
\midrule
LTX-2 &1.05 & 116.24& \textbf{0.08} & 0.56 & \textbf{1.81}& \textbf{2.45} & \textbf{5.82} & \textbf{4.92} & 5.51 & \textbf{6.21} & \textbf{0.25} & 0.05 & \textbf{0.92} & \textbf{1.00} & 0.89 \\
\midrule
Mova & 0.21& 86.09& -0.26 & 0.31 & 1.51& 1.56 & 4.36 & 4.80 & 5.25 & 5.87 & 0.20 & 0.08 & 0.80 & 0.98 & 0.71 \\
\midrule
Ovi &1.27 & 93.37& -0.09 & \textbf{1.20} & 1.40 & 2.40 & 5.50 & 4.86 & \textbf{5.63} & 5.99 & 0.23 & \textbf{0.04} & 0.91 & \textbf{1.00} & \textbf{0.95} \\
\midrule
Ours & \textbf{20.38} & \textbf{0.94}& -0.16 & 1.12 & 1.37 & 2.32 & 4.72 & 4.65 & 5.16 & 5.53 & 0.21 & 0.09 & 0.90 & 0.98 & 0.92 \\
\bottomrule
\end{tabular}
}
\vspace{3pt}
\caption{Quantitative evaluation on the Text-to-Audio-Video (T2AV) task.}
\label{tab:t2av_metrics_t2av}
\end{table*}
Training is conducted on 16 GPUs with Fully Sharded Data Parallel (FSDP), using a global batch size of 16 and a learning rate of $2 \times 10^{-6}$. 
Our two-stage optimization follows the training pipeline described in Sec.~3.4: Stage I (Dual-Stream ODE Initialization) runs for 3,000 steps, and Stage II (Self-Rollout and Dual-Stream DMD) runs for 2,000 steps. 

For the data pipeline, we begin with 100 seed prompts written by a human annotator and expand them through prompt rewriting and paraphrasing using Qwen3.5-Plus \cite{team2026qwen3} to obtain a substantially larger prompt pool.
We then use the pretrained Ovi model to generate corresponding audio-video samples and perform prompt-level filtering based on generation quality indicators, including Sync Confidence score \cite{chung2016out} and WER, retaining prompts that yield stable synchronization and reliable speech content. Further details are provided in Appendix~\ref{sec:data_construction}.

\noindentparagraph{Evaluation Metrics.}
We evaluate Hallo-Live from six complementary perspectives: real-time efficiency, video fidelity, audio-visual synchronization, acoustic naturalness, TTS-oriented speech-text consistency, and human-centric portrait fidelity. For efficiency, we report throughput (FPS) and latency (s), both measured on two NVIDIA H200 GPUs. For visual quality, we adopt VideoAlign \cite{liu2025improving}, including Visual Quality (VQ), Motion Quality (MQ), Text-Alignment (TA), and the Overall score. For cross-modal alignment, we use SyncNet \cite{chung2016out} confidence to measure the correspondence between lip motion and speech. For audio quality, we follow AudioBox \cite{tjandra2025meta} and report Content Enjoyment (CE), Content Usefulness (CU), and Production Quality (PQ). For TTS-oriented evaluation, we additionally report CLAP score and word error rate (WER), where a higher CLAP and lower WER indicate better text-audio alignment and speech intelligibility. Finally, to better capture avatar-specific artifacts, we additionally report Human Fidelity on Anatomy (Anat.), Clothing (Clo.), and Identity (Id.) in VBench-2.0 \cite{zheng2025vbench}.

\subsection{Comparison results}

We compare Hallo-Live with representative joint audio-visual generation frameworks, including JavisDiT \cite{liu2025javisdit}, UniVerse-1 \cite{wang2025universe}, Ovi \cite{low2025ovi}, MOVA \cite{team2026mova}, and LTX-2 \cite{hacohen2026ltx}. We do not include OmniForcing \cite{su2026omniforcing} in the comparison because its checkpoints are not publicly available. Quantitative results are summarized in Table~\ref{tab:t2av_metrics_t2av}, and qualitative comparisons are shown in Figure~\ref{fig:main_results}. A clear trend emerges: Hallo-Live is the only method that reaches the real-time regime while preserving generation quality close to the much heavier Ovi teacher.

\noindentparagraph{Analysis of Inference Efficiency}The most significant advantage of Hallo-Live is inference efficiency. Our model reaches 20.38 FPS with only 0.94 seconds of latency on two H200 GPUs, whereas all baselines remain below 2.15 FPS and require at least 24.40 seconds before generation begins. Relative to the Ovi teacher, Hallo-Live improves throughput by about $16.0\times$ (20.38 vs.\ 1.27 FPS) and reduces latency by about $99.3\times$ (0.94 vs.\ 93.37 seconds). This gap is large enough to change the deployment setting: previous systems are primarily suitable for offline generation, while Hallo-Live is practical for responsive avatar interaction.
\noindentparagraph{Analysis of Generation Quality}Despite this aggressive acceleration, Hallo-Live preserves strong generation quality. On VideoAlign, our method achieves an overall score of 2.32, only 0.08 lower than Ovi (2.40) and 0.13 lower than LTX-2 (2.45), while substantially outperforming JavisDiT, UniVerse-1, and MOVA. Human-centric portrait fidelity is also well maintained: Hallo-Live obtains 0.90 on anatomy, 0.98 on clothing, and 0.92 on identity consistency, nearly matching Ovi (0.91/1.00/0.95). These results indicate that the proposed asynchronous dual-stream design retains most of the teacher's visual prior under streaming inference.

Across synchronization, speech quality, and TTS-oriented metrics, Hallo-Live remains well balanced. Our method achieves a Sync score of 4.72, outperforming JavisDiT, UniVerse-1, and MOVA, while remaining below Ovi (5.50) and LTX-2 (5.82). On AudioBox, Hallo-Live remains competitive across all three reported dimensions, suggesting that the synthesized speech preserves good naturalness under streaming generation. This trend is consistent with the TTS-oriented results in Table~\ref{tab:t2av_metrics_t2av}: Hallo-Live attains a CLAP score of 0.21, comparable to JavisDiT (0.19), UniVerse-1 (0.18), and MOVA (0.20), though still below Ovi (0.23) and LTX-2 (0.25). Its WER of 0.09 is markedly better than JavisDiT (0.88) and remains reasonably close to MOVA (0.08) and UniVerse-1 (0.07), though it still trails Ovi (0.04) and LTX-2 (0.05). Although Hallo-Live is not the best standalone TTS system, these results indicate that it preserves good text-audio alignment and intelligibility while prioritizing real-time joint audio-video generation.

\begin{table}[tb]
  \centering
  \small 
  \setlength{\tabcolsep}{3pt} 
  \renewcommand{\arraystretch}{1.1} 
  \resizebox{\linewidth}{!}{%
  \begin{tabular}{p{3.3cm} c c c c}
    \toprule
    Attention mechanism & Window size & Sync-C $\uparrow$  & VideoAlign (Overall) $\uparrow$&AudioBox (Avg.)$\uparrow$\\
    \midrule
    Strict Block-Causal Attention  & -& 3.87& 2.09& 5.11 \\
    \midrule
    Future-Expanding Attention  & 5 & 4.08& 1.98& \textbf{5.13}\\
    \midrule
    Future-Expanding Attention & 10 & 4.22&\textbf{2.16} & 5.07 \\
    \midrule
    Future-Expanding Attention  & 15 & 4.29& 1.97& 5.03 \\
    \midrule
    Future-Expanding Attention  & 30 & \textbf{4.33}& 2.03& 4.95  \\
    \bottomrule
  \end{tabular}
  }
  \vspace{3pt}
  \caption{Ablation study of different attention mechanisms}
  \label{tab:ablation_attention}
\end{table}

\begin{figure}[tb]
    \centering
    \includegraphics[width=\linewidth]{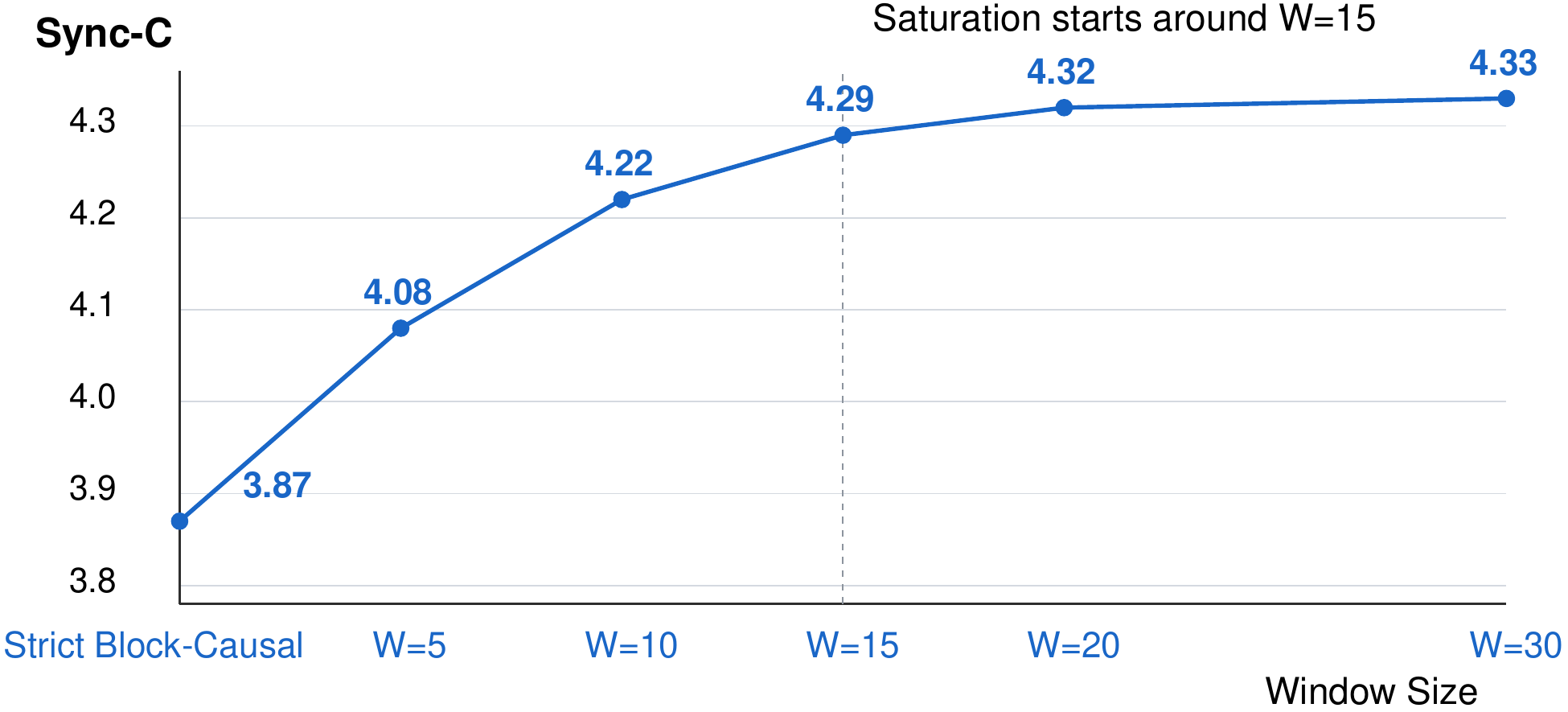}
    \caption{Line plot of the Sync-C score under different attention mechanisms. The score increases steadily as the attention window expands, while the improvement becomes marginal after $W=15$, indicating a clear saturation trend.}
    \label{fig:ablation_attention}
\end{figure}

The qualitative results in Figure~\ref{fig:main_results} further support these findings. Hallo-Live produces more visually stable portraits, cleaner identity preservation, and more coherent lip motion than other efficient baselines. Figure~\ref{fig:prompt} additionally shows that the model generalizes well across diverse prompt conditions, including half-body and full-body compositions, multi-speaker scenes, and cartoon-style synthesis. Overall, these results demonstrate that Hallo-Live offers the strongest quality-efficiency trade-off among the compared text-to-audio-video generation systems.

\subsection{Ablation results}
\noindentparagraph{Different Attention Mechanisms}
\begin{figure*}[t]
    \centering
    \includegraphics[width=\textwidth]{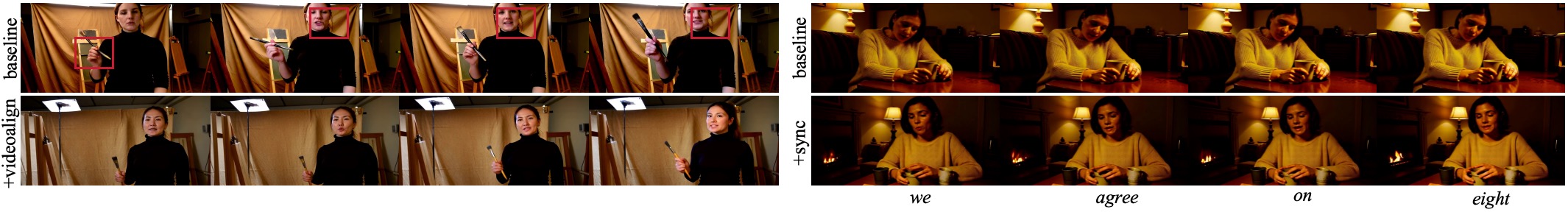}
    \caption{Qualitative comparison of individual reward enhancements. The reward-weighted distillation allows the student to pull its distribution toward high-reward regions. Each row highlights improvements in lip-sync precision (bottom)and video aesthetic quality (top) compared to the standard DMD baseline.}
    \label{fig:compare_base_rl}
\end{figure*}

Table~\ref{tab:ablation_attention} and Figure~\ref{fig:ablation_attention} compare the original strict block-causal attention with our Future-Expanding Attention under different window sizes. Replacing the strict block-causal pattern with the proposed future-expanding window consistently improves audio-video synchronization: the Sync Confidence score increases from 3.87 to 4.08, 4.22, 4.29, and 4.33 as $W$ grows from 5 to 30. This trend verifies the core motivation of our design: allowing the video stream to access a short horizon of future audio cues is critical for modeling anticipatory lip motion. At the same time, Figure~\ref{fig:ablation_attention} reveals a clear saturation effect, where the gain is substantial from the block-causal baseline to moderate window sizes, but becomes marginal after $W{=}15$. This suggests that most useful phonetic context is concentrated within a limited temporal range, and simply enlarging the receptive field brings diminishing synchronization returns.


\begin{table}[tb]
  \centering
  \small 
  \setlength{\tabcolsep}{1pt} 
  \renewcommand{\arraystretch}{1.1} 
  \begin{tabular}{l cccc c ccc}
\toprule
\multirow{2}{*}{Method} & \multicolumn{4}{c}{VideoAlign} & \multirow{2}{*}{Sync $\uparrow$} & \multicolumn{3}{c}{AudioBox}  \\
\cmidrule(lr){2-5} \cmidrule(lr){7-9}
 & VQ $\uparrow$ & MQ $\uparrow$ & TA $\uparrow$ & Overall $\uparrow$ & & CE $\uparrow$ & CU $\uparrow$ & PQ $\uparrow$  \\
\midrule
baseline & -0.35 & 1.00 & \textbf{1.38} & 2.03  & 4.33 & 4.65 & 5.04 & 5.45 \\
\midrule
+VideoAlign      & \textbf{-0.12} & \textbf{1.14} & 1.32 & \textbf{2.34} & 3.93 & 4.59 & 5.08 & 5.53  \\
\midrule
+Sync            & -0.29 & 0.96 & 1.37 & 2.04 & \textbf{5.37} & 4.64 & 5.20 & 5.63  \\
\midrule
+AudioBox        & -0.15 & 0.97 & 1.28 & 2.10 & 4.03 & \textbf{4.75} &\textbf{5.27} & \textbf{5.88}  \\
\midrule
+All             & -0.16 & 1.12 & 1.37 & 2.32 & 4.72 & 4.65 & 5.16 & 5.53  \\
\bottomrule
\end{tabular}
\vspace{3pt}
\caption{Ablation of multi-modal preference guidance with individual and joint rewards.}
  \label{tab:t2av_metrics_ablation_rl}
\end{table}


\begin{table}[tb]
  \centering
  \small 
  \setlength{\tabcolsep}{1pt} 
  \renewcommand{\arraystretch}{1.1} 
  \begin{tabular}{l cccc c ccc}
\toprule
\multirow{2}{*}{configuration} & \multicolumn{4}{c}{VideoAlign} & \multirow{2}{*}{Sync $\uparrow$} & \multicolumn{3}{c}{AudioBox}  \\
\cmidrule(lr){2-5} \cmidrule(lr){7-9}
 & VQ $\uparrow$ & MQ $\uparrow$ & TA $\uparrow$ & Overall $\uparrow$ & & CE $\uparrow$ & CU $\uparrow$ & PQ $\uparrow$  \\
\midrule
baseline & -0.35 & 1.00 & \textbf{1.38} & 2.03  & 4.33 & \textbf{4.65} & 5.04 & 5.45 \\
\midrule
+Sync,$\beta$=1      & -0.29 &\textbf{1.10}& 1.19 & 2.00& 4.58 & 4.60 &5.10 & 5.49 \\
\midrule
+Sync,$\beta$=2      & -0.29 & 0.96 & 1.37 & \textbf{2.04} & \textbf{5.37} & 4.64 & \textbf{5.20} & \textbf{5.63}  \\
\midrule
+Sync,$\beta$=3      & \textbf{-0.15} &0.56  & 1.21 &1.62 & 3.30 & 4.56 &  4.91& 5.40  \\
\midrule
+Sync,$\beta$=4      & -0.47&0.46  & 1.12 & 1.23& 3.11& 4.33 & 4.77 &   5.24\\
\bottomrule
\end{tabular}
\vspace{3pt}
\caption{Ablation of the reward coefficient $\beta$ under Sync-only reward weighting.}
  \label{tab:ablation_metrics_weight}
\end{table}

\begin{table}[tb]
  \centering
  \small 
  \setlength{\tabcolsep}{1pt} 
  \renewcommand{\arraystretch}{1.1} 
  \begin{tabular}{l cccc c ccc}
\toprule
\multirow{2}{*}{configuration} & \multicolumn{4}{c}{VideoAlign} & \multirow{2}{*}{Sync $\uparrow$} & \multicolumn{3}{c}{AudioBox}  \\
\cmidrule(lr){2-5} \cmidrule(lr){7-9}
 & VQ $\uparrow$ & MQ $\uparrow$ & TA $\uparrow$ & Overall $\uparrow$ & & CE $\uparrow$ & CU $\uparrow$ & PQ $\uparrow$  \\
\midrule
baseline & -0.35 & 1.00 & \textbf{1.38} & 2.03  &\textbf{4.33} & \textbf{4.65} & 5.04 & 5.45 \\
\midrule
+VideoAlign,$\beta$=1      & -0.24 &1.05& 1.35 & 2.11 & 4.10 &4.56  &5.01  &  \textbf{5.56} \\
\midrule
+VideoAlign,$\beta$=2      & \textbf{-0.12} & \textbf{1.14} & 1.32 & \textbf{2.34} & 3.93 & 4.59 & \textbf{5.08} & 5.53  \\
\midrule
+VideoAlign,$\beta$=3        & -0.34 & 0.77 &1.22  & 1.85& 3.19 & 4.58 &4.95 & 5.39 \\
\midrule
+VideoAlign,$\beta$=4             & -0.44 & 0.43 & 1.16 &1.25 & 3.03& 4.43 &4.92  & 5.35  \\
\bottomrule
\end{tabular}
\vspace{3pt}
\caption{Ablation of the reward coefficient $\beta$ under VideoAlign-only reward weighting.}
  \label{tab:1ablation_rl}
\end{table}

\noindentparagraph{Effectiveness of Multi-Modal Preference Guidance}
To evaluate the proposed Human-Centric Preference-Guided DMD, we start from the distilled streaming student without reward weighting and then add the VideoAlign, Sync, and AudioBox rewards individually and jointly. Table~\ref{tab:t2av_metrics_ablation_rl} shows that each reward primarily improves the modality it explicitly supervises. Adding the VideoAlign reward yields the strongest visual gains, improving from $-0.35/1.00/2.03$ to $-0.12/1.14/2.34$. Adding the Sync reward produces the largest improvement in audio-visual alignment, increasing the Sync score from 4.33 to 5.37. Adding the AudioBox reward most effectively enhances acoustic quality, achieving the best CE/CU/PQ scores of 4.75/5.27/5.88 among all single-reward settings.

These results reveal a clear pattern: single-reward optimization is highly targeted, but its benefits transfer only weakly to the other modalities. In particular, the Sync-only setting substantially improves synchronization, while its effect on VideoAlign and AudioBox remains limited, indicating that synchronization reward alone is insufficient to ensure balanced visual and acoustic quality. By contrast, jointly combining all three rewards yields the most balanced trade-off, achieving strong visual quality (2.32 VideoAlign) and  reliable synchronization (4.72 Sync). Figure~\ref{fig:compare_base_rl} further confirms this trend qualitatively: compared with vanilla DMD, multi-modal reward guidance produces sharper visual details and more accurate lip-audio alignment. These observations validate the necessity of jointly constraining synchronization, visual fidelity, and audio naturalness in preference-guided distillation.

\begin{table}[tb]
  \centering
  \small 
  \setlength{\tabcolsep}{1pt} 
  \renewcommand{\arraystretch}{1.1} 
  \begin{tabular}{l cccc c ccc}
\toprule
\multirow{2}{*}{configuration} & \multicolumn{4}{c}{VideoAlign} & \multirow{2}{*}{Sync $\uparrow$} & \multicolumn{3}{c}{AudioBox}  \\
\cmidrule(lr){2-5} \cmidrule(lr){7-9}
 & VQ $\uparrow$ & MQ $\uparrow$ & TA $\uparrow$ & Overall $\uparrow$ & & CE $\uparrow$ & CU $\uparrow$ & PQ $\uparrow$  \\
\midrule
baseline & -0.35 & 1.00 & \textbf{1.38} & 2.03  & \textbf{4.33} & 4.65 & 5.04 & 5.45 \\
\midrule
+AudioBox,$\beta$=1      &-0.29  &  \textbf{1.05}&1.23 & 2.05 & 4.09 & 4.69& 5.17 &5.65   \\
\midrule
+AudioBox,$\beta$=2        & \textbf{-0.15} & 1.03 & 1.28 &\textbf{2.10} & 4.03 & \textbf{4.75} &\textbf{5.27} & \textbf{5.88}  \\
\midrule
+AudioBox,$\beta$=3        &-0.37 &0.87  &1.21 & 1.76& 3.25 &4.59  &4.97 &  5.47 \\
\midrule
+AudioBox,$\beta$=4        & -0.49 &0.45  & 1.12 & 1.16& 3.08 & 4.54 & 4.96 &5.37  \\
\bottomrule
\end{tabular}
\vspace{3pt}
\caption{Ablation of the reward coefficient $\beta$ under AudioBox-only reward weighting.}
  \label{tab:2ablation_rl}
\end{table}

\noindentparagraph{Reward Coefficients Ablation}

We conduct a systematic investigation into the sensitivity of the reward coefficient $\beta$ across three distinct preference settings: \emph{Sync-only}, \emph{VideoAlign-only}, and \emph{AudioBox-only}. This analysis, supported by quantitative results in Tables \ref{tab:ablation_metrics_weight}, \ref{tab:1ablation_rl}, \ref{tab:2ablation_rl} , aims to elucidate the mechanism of our reward-weighted objective and its impact on multimodal alignment.

Empirical evidence suggests that the model's performance is highly sensitive to the choice of $\beta$. In the \emph{Sync-only} setting, increasing $\beta$ from 1 to 2 yields a marked improvement in the Sync score (from 4.58 to 5.37), indicating that a moderate reward weight is essential for the student model to effectively internalize the synchronization signals. This trend is consistently observed across the \emph{VideoAlign-only} and \emph{AudioBox-only} experiments (Tables \ref{tab:1ablation_rl} and \ref{tab:2ablation_rl}), where $\beta=2$ emerges as a universal "sweet spot." At this configuration, the model achieves the optimal trade-off between various multimodal alignment metrics and generation fidelity.

Conversely, we observe a sharp deterioration in all metrics when $\beta$ exceeds the threshold of 2. For instance, in the VideoAlign-only setting, increasing $\beta$ to 4 reduces the VideoAlign Overall score to 1.25. In the Sync-only setting, the Sync score drops from 5.37 at $\beta=2$ to 3.30 and 3.11 at $\beta=3$ and $\beta=4$, respectively. We attribute this performance collapse to reward hacking: an excessively high coefficient over-amplifies the reward signal, driving the model into pathological regions of the latent space that yield high rewards but severely compromise generation stability and overall quality.

Based on these ablation studies, $\beta=2$ serves as the critical hyperparameter that maximizes alignment performance while avoiding the pitfalls of over-optimization. Consequently, we adopt $\beta=2$ as the default configuration for our final method.

\section{Conclusion}
In this paper, we introduced \textbf{Hallo-Live}, a real-time framework for text-driven joint audio-video avatar generation. By combining Asynchronous Dual-Stream and Human-Centric Preference-Guided DMD, the model improves lip-audio synchronization under streaming inference while reducing the quality loss caused by aggressive acceleration.

Experiments show that Hallo-Live achieves 20.38 FPS with 0.94 seconds latency on two NVIDIA H200 GPUs, giving $16.0\times$ higher throughput and $99.3\times$ lower latency than the Ovi teacher while maintaining strong VideoAlign, Sync, and human-centric fidelity results. The model also generalizes well across diverse prompt conditions, including photorealistic portraits, multi-speaker scenes, and stylized avatars. These results suggest that Hallo-Live is a practical step toward deployable interactive avatar generation. Future work will explore longer-horizon conversations, richer body and camera control, and deployment on lower-cost hardware.


{
    \bibliographystyle{ACM-Reference-Format}
    \bibliography{references}

@inproceedings{yin2024one,
  title     = {One-step diffusion with distribution matching distillation},
  author    = {Yin, Tianwei and Gharbi, Micha{\"e}l and Zhang, Richard and Shechtman, Eli and Durand, Fredo and Freeman, William T and Park, Taesung},
  booktitle = {Proceedings of the IEEE/CVF conference on computer vision and pattern recognition},
  pages     = {6613--6623},
  year      = {2024}
}

@article{yin2024improved,
  title   = {Improved distribution matching distillation for fast image synthesis},
  author  = {Yin, Tianwei and Gharbi, Micha{\"e}l and Park, Taesung and Zhang, Richard and Shechtman, Eli and Durand, Fredo and Freeman, William T},
  journal = {Advances in neural information processing systems},
  volume  = {37},
  pages   = {47455--47487},
  year    = {2024}
}

@article{jiang2025distribution,
  title   = {{Distribution Matching Distillation Meets Reinforcement Learning}},
  author  = {Jiang, Dengyang and Liu, Dongyang and Wang, Zanyi and Wu, Qilong and Jin, Xin and Liu, David and Li, Zhen and Wang, Mengmeng and Gao, Peng and Yang, Harry},
  journal = {arXiv preprint arXiv:2511.13649},
  year    = {2025}
}

@article{luo2025learning,
  title   = {Learning Few-Step Diffusion Models by Trajectory Distribution Matching},
  author  = {Luo, Yihong and Hu, Tianyang and Sun, Jiacheng and Cai, Yujun and Tang, Jing},
  journal = {arXiv preprint arXiv:2503.06674},
  year    = {2025}
}

@article{xu2025one,
  title   = {One-step Diffusion Models with {$f$}-Divergence Distribution Matching},
  author  = {Xu, Yilun and Nie, Weili and Vahdat, Arash},
  journal = {arXiv preprint arXiv:2502.15681},
  year    = {2025}
}

@inproceedings{prajwal2020wav2lip,
  title     = {A Lip Sync Expert Is All You Need for Speech to Lip Generation In The Wild},
  author    = {Prajwal, K R and Mukhopadhyay, Rudrabha and Namboodiri, Vinay and Jawahar, C V},
  booktitle = {Proceedings of the 28th ACM International Conference on Multimedia},
  pages     = {484--492},
  year      = {2020}
}

@inproceedings{zhang2023sadtalker,
  title     = {SadTalker: Learning Realistic 3D Motion Coefficients for Stylized Audio-Driven Single Image Talking Face Animation},
  author    = {Zhang, Wenxuan and Cun, Xiaodong and Wang, Xuan and Zhang, Yong and Shen, Xi and Guo, Yu and Shan, Ying and Wang, Fei},
  booktitle = {Proceedings of the IEEE/CVF Conference on Computer Vision and Pattern Recognition},
  pages     = {8652--8661},
  year      = {2023}
}

@article{tian2024emo,
  title   = {{EMO}: Emote Portrait Alive -- Generating Expressive Portrait Videos with Audio2Video Diffusion Model under Weak Conditions},
  author  = {Tian, Linrui and Wang, Qi and Zhang, Bang and Bo, Liefeng},
  journal = {arXiv preprint arXiv:2402.17485},
  year    = {2024}
}

@article{xu2024vasa1,
  title   = {{VASA-1}: Lifelike Audio-Driven Talking Faces Generated in Real Time},
  author  = {Xu, Sicheng and Chen, Guojun and Guo, Yu-Xiao and Yang, Jiaolong and Li, Chong and Zang, Zhenyu and Zhang, Yizhong and Tong, Xin and Guo, Baining},
  journal = {arXiv preprint arXiv:2404.10667},
  year    = {2024}
}

@article{xu2024hallo,
  title   = {{Hallo}: Hierarchical Audio-Driven Visual Synthesis for Portrait Image Animation},
  author  = {Xu, Mingwang and Li, Hui and Su, Qingkun and Shang, Hanlin and Zhang, Liwei and Liu, Ce and Wang, Jingdong and Yao, Yao and Zhu, Siyu},
  journal = {arXiv preprint arXiv:2406.08801},
  year    = {2024}
}

@article{cui2024hallo2,
  title   = {{Hallo2}: Long-Duration and High-Resolution Audio-Driven Portrait Image Animation},
  author  = {Cui, Jiahao and Li, Hui and Yao, Yao and Zhu, Hao and Shang, Hanlin and Cheng, Kaihui and Zhou, Hang and Zhu, Siyu and Wang, Jingdong},
  journal = {arXiv preprint arXiv:2410.07718},
  year    = {2024}
}

@article{zhen2025teller,
  title   = {{Teller}: Real-Time Streaming Audio-Driven Portrait Animation with Autoregressive Motion Generation},
  author  = {Zhen, Dingcheng and Yin, Shunshun and Qin, Shiyang and Yi, Hou and Zhang, Ziwei and Liu, Siyuan and Qi, Gan and Tao, Ming},
  journal = {arXiv preprint arXiv:2503.18429},
  year    = {2025}
}

@article{gan2025omniavatar,
  title   = {{OmniAvatar}: Efficient Audio-Driven Avatar Video Generation with Adaptive Body Animation},
  author  = {Gan, Qijun and Yang, Ruizi and Zhu, Jianke and Xue, Shaofei and Hoi, Steven},
  journal = {arXiv preprint arXiv:2506.18866},
  year    = {2025}
}

@article{liu2025javisdit,
  title   = {{JavisDiT}: Joint Audio-Video Diffusion Transformer with Hierarchical Spatio-Temporal Prior Synchronization},
  author  = {Liu, Kai and Li, Wei and Chen, Lai and Wu, Shengqiong and Zheng, Yanhao and Ji, Jiayi and Zhou, Fan and Jiang, Rongxin and Luo, Jiebo and Fei, Hao and Chua, Tat-Seng},
  journal = {arXiv preprint arXiv:2503.23377},
  year    = {2025}
}

@article{low2025ovi,
  title   = {Ovi: Twin backbone cross-modal fusion for audio-video generation},
  author  = {Low, Chetwin and Wang, Weimin and Katyal, Calder},
  journal = {arXiv preprint arXiv:2510.01284},
  year    = {2025}
}

@inproceedings{ronneberger2015u,
  title        = {U-net: Convolutional networks for biomedical image segmentation},
  author       = {Ronneberger, Olaf and Fischer, Philipp and Brox, Thomas},
  booktitle    = {International Conference on Medical image computing and computer-assisted intervention},
  pages        = {234--241},
  year         = {2015},
  organization = {Springer}
}

@article{wang2025universe,
  title   = {UniVerse-1: Unified Audio-Video Generation via Stitching of Experts},
  author  = {Wang, Duomin and Zuo, Wei and Li, Aojie and Chen, Ling-Hao and Liao, Xinyao and Zhou, Deyu and Yin, Zixin and Dai, Xili and Jiang, Daxin and Yu, Gang},
  journal = {arXiv preprint arXiv:2509.06155},
  year    = {2025}
}

@article{shazeer2017outrageously,
  title   = {Outrageously large neural networks: The sparsely-gated mixture-of-experts layer},
  author  = {Shazeer, Noam and Mirhoseini, Azalia and Maziarz, Krzysztof and Davis, Andy and Le, Quoc and Hinton, Geoffrey and Dean, Jeff},
  journal = {arXiv preprint arXiv:1701.06538},
  year    = {2017}
}

@article{team2026mova,
  title   = {Mova: Towards scalable and synchronized video-audio generation},
  author  = {Team, OpenMOSS and Yu, Donghua and Chen, Mingshu and Chen, Qi and Luo, Qi and Wu, Qianyi and Cheng, Qinyuan and Li, Ruixiao and Liang, Tianyi and Zhang, Wenbo and others},
  journal = {arXiv preprint arXiv:2602.08794},
  year    = {2026}
}

@article{chern2026speed,
  title   = {Speed by Simplicity: A Single-Stream Architecture for Fast Audio-Video Generative Foundation Model},
  author  = {Chern, Ethan and Teng, Hansi and Sun, Hanwen and Wang, Hao and Pan, Hong and Jia, Hongyu and Su, Jiadi and Li, Jin and Yu, Junjie and Liu, Lijie and others},
  journal = {arXiv preprint arXiv:2603.21986},
  year    = {2026}
}

@article{zhang2026foleycrafter,
  title     = {Foleycrafter: Bring silent videos to life with lifelike and synchronized sounds},
  author    = {Zhang, Yiming and Gu, Yicheng and Zeng, Yanhong and Xing, Zhening and Wang, Yuancheng and Wu, Zhizheng and Liu, Bin and Chen, Kai},
  journal   = {International Journal of Computer Vision},
  volume    = {134},
  number    = {1},
  pages     = {46},
  year      = {2026},
  publisher = {Springer}
}

@article{hacohen2026ltx,
  title   = {LTX-2: Efficient Joint Audio-Visual Foundation Model},
  author  = {HaCohen, Yoav and Brazowski, Benny and Chiprut, Nisan and Bitterman, Yaki and Kvochko, Andrew and Berkowitz, Avishai and Shalem, Daniel and Lifschitz, Daphna and Moshe, Dudu and Porat, Eitan and others},
  journal = {arXiv preprint arXiv:2601.03233},
  year    = {2026}
}

@article{su2026omniforcing,
  title   = {{OmniForcing}: Unleashing Real-time Joint Audio-Visual Generation},
  author  = {Su, Yaofeng and Li, Yuming and Xue, Zeyue and Huang, Jie and Fu, Siming and Li, Haoran and Li, Ying and Qian, Zezhong and Huang, Haoyang and Duan, Nan},
  journal = {arXiv preprint arXiv:2603.11647},
  year    = {2026}
}

@article{black2023ddpo,
  title   = {Training Diffusion Models with Reinforcement Learning},
  author  = {Black, Kevin and Janner, Michael and Du, Yilun and Kostrikov, Ilya and Levine, Sergey},
  journal = {arXiv preprint arXiv:2305.13301},
  year    = {2023}
}

@article{wan2025wan,
  title   = {Wan: Open and advanced large-scale video generative models},
  author  = {Wan, Team and Wang, Ang and Ai, Baole and Wen, Bin and Mao, Chaojie and Xie, Chen-Wei and Chen, Di and Yu, Feiwu and Zhao, Haiming and Yang, Jianxiao and others},
  journal = {arXiv preprint arXiv:2503.20314},
  year    = {2025}
}

@inproceedings{peebles2023scalable,
  title     = {Scalable diffusion models with transformers},
  author    = {Peebles, William and Xie, Saining},
  booktitle = {Proceedings of the IEEE/CVF international conference on computer vision},
  pages     = {4195--4205},
  year      = {2023}
}

@article{vaswani2017attention,
  title   = {Attention is all you need},
  author  = {Vaswani, Ashish and Shazeer, Noam and Parmar, Niki and Uszkoreit, Jakob and Jones, Llion and Gomez, Aidan N and Kaiser, {\L}ukasz and Polosukhin, Illia},
  journal = {Advances in neural information processing systems},
  volume  = {30},
  year    = {2017}
}

@article{su2024roformer,
  title     = {Roformer: Enhanced transformer with rotary position embedding},
  author    = {Su, Jianlin and Ahmed, Murtadha and Lu, Yu and Pan, Shengfeng and Bo, Wen and Liu, Yunfeng},
  journal   = {Neurocomputing},
  volume    = {568},
  pages     = {127063},
  year      = {2024},
  publisher = {Elsevier}
}

@inproceedings{rombach2022high,
  title     = {High-resolution image synthesis with latent diffusion models},
  author    = {Rombach, Robin and Blattmann, Andreas and Lorenz, Dominik and Esser, Patrick and Ommer, Bj{\"o}rn},
  booktitle = {Proceedings of the IEEE/CVF conference on computer vision and pattern recognition},
  pages     = {10684--10695},
  year      = {2022}
}

@inproceedings{ruan2023mm,
  title     = {Mm-diffusion: Learning multi-modal diffusion models for joint audio and video generation},
  author    = {Ruan, Ludan and Ma, Yiyang and Yang, Huan and He, Huiguo and Liu, Bei and Fu, Jianlong and Yuan, Nicholas Jing and Jin, Qin and Guo, Baining},
  booktitle = {Proceedings of the IEEE/CVF conference on computer vision and pattern recognition},
  pages     = {10219--10228},
  year      = {2023}
}

@article{lee2023aligning,
  title   = {Aligning Text-to-Image Models using Human Feedback},
  author  = {Lee, Kimin and Liu, Hao and Ryu, Moonkyung and Watkins, Olivia and Du, Yuqing and Boutilier, Craig and Abbeel, Pieter and Ghavamzadeh, Mohammad and Gu, Shixiang Shane},
  journal = {arXiv preprint arXiv:2302.12192},
  year    = {2023}
}

@article{liu2025improving,
  title   = {Improving video generation with human feedback},
  author  = {Liu, Jie and Liu, Gongye and Liang, Jiajun and Yuan, Ziyang and Liu, Xiaokun and Zheng, Mingwu and Wu, Xiele and Wang, Qiulin and Xia, Menghan and Wang, Xintao and others},
  journal = {arXiv preprint arXiv:2501.13918},
  year    = {2025}
}

@inproceedings{cui2025hallo3,
  title     = {Hallo3: Highly dynamic and realistic portrait image animation with video diffusion transformer},
  author    = {Cui, Jiahao and Li, Hui and Zhan, Yun and Shang, Hanlin and Cheng, Kaihui and Ma, Yuqi and Mu, Shan and Zhou, Hang and Wang, Jingdong and Zhu, Siyu},
  booktitle = {Proceedings of the Computer Vision and Pattern Recognition Conference},
  pages     = {21086--21095},
  year      = {2025}
}

@article{cui2025hallo4,
  title   = {Hallo4: High-fidelity dynamic portrait animation via direct preference optimization and temporal motion modulation},
  author  = {Cui, Jiahao and Chen, Yan and Xu, Mingwang and Shang, Hanlin and Chen, Yuxuan and Zhan, Yun and Dong, Zilong and Yao, Yao and Wang, Jingdong and Zhu, Siyu},
  journal = {arXiv e-prints},
  pages   = {arXiv--2505},
  year    = {2025}
}

@article{li2024latentsync,
  title   = {Latentsync: Taming audio-conditioned latent diffusion models for lip sync with syncnet supervision},
  author  = {Li, Chunyu and Zhang, Chao and Xu, Weikai and Lin, Jingyu and Xie, Jinghui and Feng, Weiguo and Peng, Bingyue and Chen, Cunjian and Xing, Weiwei},
  journal = {arXiv preprint arXiv:2412.09262},
  year    = {2024}
}

@article{tjandra2025meta,
  title   = {Meta audiobox aesthetics: Unified automatic quality assessment for speech, music, and sound},
  author  = {Tjandra, Andros and Wu, Yi-Chiao and Guo, Baishan and Hoffman, John and Ellis, Brian and Vyas, Apoorv and Shi, Bowen and Chen, Sanyuan and Le, Matt and Zacharov, Nick and others},
  journal = {arXiv preprint arXiv:2502.05139},
  year    = {2025}
}

@article{team2026qwen3,
  title   = {Qwen3. 5-Omni Technical Report},
  author  = {Team, Qwen},
  journal = {arXiv preprint arXiv:2604.15804},
  year    = {2026}
}

@inproceedings{chung2016out,
  title        = {Out of time: automated lip sync in the wild},
  author       = {Chung, Joon Son and Zisserman, Andrew},
  booktitle    = {Asian conference on computer vision},
  pages        = {251--263},
  year         = {2016},
  organization = {Springer}
}

@article{2020t5,
  author  = {Colin Raffel and Noam Shazeer and Adam Roberts and Katherine Lee and Sharan Narang and Michael Matena and Yanqi Zhou and Wei Li and Peter J. Liu},
  title   = {Exploring the Limits of Transfer Learning with a Unified Text-to-Text Transformer},
  journal = {Journal of Machine Learning Research},
  year    = {2020},
  volume  = {21},
  number  = {140},
  pages   = {1-67},
  url     = {http://jmlr.org/papers/v21/20-074.html}
}

@article{wang2024emu3,
  title   = {Emu3: Next-Token Prediction is All You Need},
  author  = {Wang, Xinlong and Zhang, Xiaosong and Luo, Zhengxiong and Sun, Quan and Cui, Yufeng and Wang, Jinsheng and Zhang, Fan and Wang, Yueze and Li, Zhen and Yu, Qiying and others},
  journal = {arXiv preprint arXiv:2409.18869},
  year    = {2024}
}

@article{huang2025self,
  title   = {Self forcing: Bridging the train-test gap in autoregressive video diffusion},
  author  = {Huang, Xun and Li, Zhengqi and He, Guande and Zhou, Mingyuan and Shechtman, Eli},
  journal = {arXiv preprint arXiv:2506.08009},
  year    = {2025}
}

@article{lu2025reward,
  title   = {Reward forcing: Efficient streaming video generation with rewarded distribution matching distillation},
  author  = {Lu, Yunhong and Zeng, Yanhong and Li, Haobo and Ouyang, Hao and Wang, Qiuyu and Cheng, Ka Leong and Zhu, Jiapeng and Cao, Hengyuan and Zhang, Zhipeng and Zhu, Xing and others},
  journal = {arXiv preprint arXiv:2512.04678},
  year    = {2025}
}

@article{ma2023dreamtalk,
  title   = {Dreamtalk: When expressive talking head generation meets diffusion probabilistic models},
  author  = {Ma, Yifeng and Zhang, Shiwei and Wang, Jiayu and Wang, Xiang and Zhang, Yingya and Deng, Zhidong},
  journal = {arXiv preprint arXiv:2312.09767},
  year    = {2023}
}

@article{chen2024echomimic,
  title   = {Echomimic: Lifelike audio-driven portrait animations through editable landmark conditions},
  author  = {Chen, Zhiyuan and Cao, Jiajiong and Chen, Zhiquan and Li, Yuming and Ma, Chenguang},
  journal = {arXiv preprint arXiv:2407.08136},
  year    = {2024}
}

@inproceedings{zhu2025infp,
  title     = {INFP: Audio-driven interactive head generation in dyadic conversations},
  author    = {Zhu, Yongming and Zhang, Longhao and Rong, Zhengkun and Hu, Tianshu and Liang, Shuang and Ge, Zhipeng},
  booktitle = {Proceedings of the IEEE/CVF Conference on Computer Vision and Pattern Recognition},
  pages     = {10667--10677},
  year      = {2025}
}

@inproceedings{jiang2024loopy,
  title     = {Loopy: Taming audio-driven portrait avatar with long-term motion dependency},
  author    = {Jiang, Jianwen and Liang, Chao and Yang, Jiaqi and Lin, Gaojie and Zhong, Tianyun and Zheng, Yanbo},
  booktitle = {The Thirteenth International Conference on Learning Representations},
  year      = {2024}
}

@article{zhang2024musetalk,
  title   = {MuseTalk: Real-Time High Quality Lip Synchronization with Latent Space Inpainting},
  author  = {Zhang, Yue and Liu, Minhao and Chen, Zhaokang and Wu, Bin and Zeng, Yubin and Zhan, Chao and He, Yingjie and Huang, Junxin and Zhou, Wenjiang},
  journal = {arXiv preprint arXiv:2410.10122},
  year    = {2024}
}

@inproceedings{mukhopadhyay2024diff2lip,
  title     = {Diff2lip: Audio conditioned diffusion models for lip-synchronization},
  author    = {Mukhopadhyay, Soumik and Suri, Saksham and Gadde, Ravi Teja and Shrivastava, Abhinav},
  booktitle = {Proceedings of the IEEE/CVF Winter Conference on Applications of Computer Vision},
  pages     = {5292--5302},
  year      = {2024}
}

@article{bigioi2024speech,
  title     = {Speech driven video editing via an audio-conditioned diffusion model},
  author    = {Bigioi, Dan and Basak, Shubhajit and Stypu{\l}kowski, Micha{\l} and Zieba, Maciej and Jordan, Hugh and McDonnell, Rachel and Corcoran, Peter},
  journal   = {Image and Vision Computing},
  volume    = {142},
  pages     = {104911},
  year      = {2024},
  publisher = {Elsevier}
}

@inproceedings{ji2025sonic,
  title     = {Sonic: Shifting focus to global audio perception in portrait animation},
  author    = {Ji, Xiaozhong and Hu, Xiaobin and Xu, Zhihong and Zhu, Junwei and Lin, Chuming and He, Qingdong and Zhang, Jiangning and Luo, Donghao and Chen, Yi and Lin, Qin and others},
  booktitle = {Proceedings of the Computer Vision and Pattern Recognition Conference},
  pages     = {193--203},
  year      = {2025}
}

@article{peng2025omnisync,
  title   = {Omnisync: Towards universal lip synchronization via diffusion transformers},
  author  = {Peng, Ziqiao and Liu, Jiwen and Zhang, Haoxian and Liu, Xiaoqiang and Tang, Songlin and Wan, Pengfei and Zhang, Di and Liu, Hongyan and He, Jun},
  journal = {arXiv preprint arXiv:2505.21448},
  year    = {2025}
}

@article{zheng2025vbench,
  title   = {Vbench-2.0: Advancing video generation benchmark suite for intrinsic faithfulness},
  author  = {Zheng, Dian and Huang, Ziqi and Liu, Hongbo and Zou, Kai and He, Yinan and Zhang, Fan and Gu, Lulu and Zhang, Yuanhan and He, Jingwen and Zheng, Wei-Shi and others},
  journal = {arXiv preprint arXiv:2503.21755},
  year    = {2025}
}
}

\clearpage
\appendix
\begin{center}
  {\LARGE\bfseries Appendix\par}
\end{center}
\thispagestyle{plain}
\section{Additional Implementation Details}

\noindentparagraph{Streaming inference procedure.}
At inference time, Hallo-Live performs block-wise streaming generation. Given a
text prompt and initial noise, the student generates the current block pair
$(V_t, A_t)$ from one video-block noise input and an expanded audio-noise input
that concatenates the current audio-block noise with extra future audio-noise
frames. This joint denoising also produces a temporary future audio block
$\tilde{A}_{t+1}$. The video branch attends only to the current visual
block, whereas the audio branch provides the expanded context
$\{\hat{A}_{t-1}, A_t, \tilde{A}_{t+1}\}$. After denoising, the committed clean
features are inserted into the rolling KV cache and the temporal window
advances to the next step. The provisional look-ahead block is never committed as
final output; instead, it is regenerated once it becomes the current block.
This overwrite strategy lets the video stream exploit short-horizon future
phonetic cues while preventing the accumulation of speculative audio errors.

\noindentparagraph{Stage II continued training.}
During Stage II dual-stream DMD training, we observe that the two streams
converge at different rates. The video stream typically stabilizes after about
2,000 optimization steps; continuing to update two streams jointly beyond this point will degrade visual quality. In
contrast, the audio stream usually requires 3,500--4,500 Stage II steps to
converge. When training is stopped before audio convergence, speech
intelligibility deteriorates substantially, as reflected by the WER of around 0.2--0.3. To balance these two optimization dynamics, we use a continued training
strategy for Stage II: the dual-stream model is first trained jointly for
2,000 steps, after which the video-stream parameters are frozen and only the
audio-stream parameters are updated for another 1,500--2,500 steps. The final
checkpoint is taken from this audio-only continued training phase, which preserves the
visual quality of the converged video stream while allowing the audio stream to reach a lower WER.

\section{Data Pipeline}
\label{sec:data_construction}
We construct the training set through a three-stage pipeline: prompt expansion, deduplication, and model-based quality filtering. We begin with 100 seed prompts written by a human annotator and expand them with Qwen3.5-Plus \cite{team2026qwen3} through prompt rewriting and paraphrasing, yielding an initial pool of 200,000 candidate prompts. We then remove near-duplicates using cosine similarity with a threshold of 0.95, resulting in 30,000 distinct prompts while preserving the semantic intent and diversity of the original seed set.

Next, we use the pretrained Ovi model to synthesize paired audio-video samples for the retained 30,000 prompts, producing approximately 42 hours of video data. To improve the reliability of the final corpus, we apply prompt-level quality filtering using a set of multimodal diagnostics. A sample is retained only if it satisfies all of the following criteria: zero word error rate (WER), VideoAlign \cite{liu2025improving} visual quality (VQ) of at least -0.8, VideoAlign text alignment (TA) of at least 0.8, Sync confidence of at least 3.0, and a VBench \cite{zheng2025vbench} human anatomy score of at least 0.7. After filtering, the final dataset contains 20,000 high-quality prompts, corresponding to approximately 28 hours of paired audio-video training data.

For clarity, the full data pipeline is summarized below:
\begin{enumerate}
    \item Expand the 100 seed prompts with Qwen3.5-Plus to obtain a large candidate prompt pool;
    \item Remove near-duplicate prompts using cosine similarity with a threshold of 0.95;
    \item Synthesize paired audio-visual samples with the pretrained Ovi model;
    \item Discard samples whose WER is non-zero;
    \item Discard samples whose VideoAlign VQ is below -0.8 or TA is below 0.8;
    \item Discard samples whose Sync Confidence score is below 3.0;
    \item Discard samples whose VBench human anatomy score is below 0.7.
\end{enumerate}

\end{document}